\documentclass[10pt,twocolumn,letterpaper]{article}
\usepackage{framed,multirow}
\usepackage{amssymb}
\usepackage{latexsym}
\usepackage{subfigure}
\usepackage{amsmath}
\usepackage{chngcntr}
\usepackage{cite}
\newcommand{\tabincell}[2]{\begin{tabular}{@{}#1@{}}#2\end{tabular}}
\usepackage{url}
\usepackage{xcolor}
\definecolor{newcolor}{rgb}{.8,.349,.1}
\usepackage[colorinlistoftodos]{todonotes}  
\usepackage{authblk}
\usepackage{geometry}
\usepackage{natbib}
\usepackage{titlesec}
\titleformat*{\section}{\large\bfseries}
\titleformat*{\subsection}{\normalsize\bfseries}

\begin{document}

\title{Faster ILOD: Incremental Learning for Object Detectors based on Faster RCNN}
\author{Can Peng}
\author{Kun Zhao}
\author{Brian C. Lovell}
\affil{School of ITEE, The University of Queensland, Brisbane, QLD, Australia}
\date{}
\maketitle

\begin{abstract}
	The human vision and perception system is inherently incremental where new knowledge is continually learned over time whilst existing knowledge is retained.
	On the other hand, deep learning networks are ill-equipped for incremental learning.
	When a well-trained network is adapted to new categories, its performance on the old categories will dramatically degrade.
	To address this problem, incremental learning methods have been explored which preserve the old knowledge of deep learning models.
	However, the state-of-the-art incremental object detector employs an external fixed region proposal method that increases overall computation time and reduces accuracy comparing to Region Proposal Network (RPN) based object detectors such as Faster RCNN.
	The purpose of this paper is to design an efficient end-to-end incremental object detector using knowledge distillation.
	We first evaluate and analyze the performance of the RPN-based detector with classic distillation on incremental detection tasks.
	Then, we introduce multi-network adaptive distillation that properly retains knowledge from the old categories when fine-tuning the model for new task.
	Experiments on the benchmark datasets, PASCAL VOC and COCO, demonstrate that the proposed incremental detector based on Faster RCNN is more accurate as well as being 13 times faster than the baseline detector.
\end{abstract}

\section{Introduction}
\label{Introduction}
\begin{figure}[tbp]
\centering
\includegraphics[width=7cm,keepaspectratio]{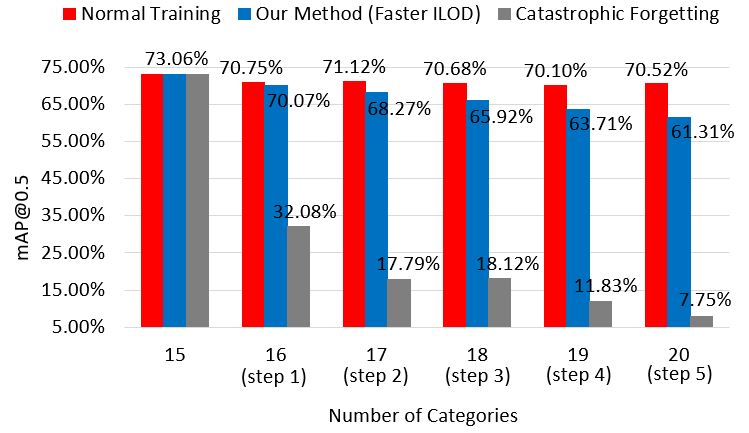}
\caption{An example of incremental object detection.
	The model is first trained on 15 categories of the PASCAL VOC dataset followed by the addition of 1 new class in each of 5 steps.
	During training, only annotations for the current learning classes are provided and all other class objects are ignored.
	Normal training retrains the model from scratch with all the available data (old and new) which gives the best possible performance.
	Catastrophic forgetting is what happens when fine-tuning the old class model with just the new class data.
	Our Faster ILOD method dramatically alleviates the forgetting problem when training on the new class data only.}
\label{fig:catastrophic forgetting}
\end{figure}

Benefiting from the rapid development of deep learning models, the performance of object detectors has increased dramatically over the years.
However, the gap between state-of-the-art performance and the human visual system is still huge.
One of the main obstacles is incrementally learning new tasks in the dynamic real-world where new categories of interest can emerge over time.
For example, in the pathology area, new sub-types of disease patterns are identified over time due to the continued growth in our knowledge and understanding.
An ideal disease pattern detection system should be able to learn a new sub-type of disease from the pathology images without losing the ability to detect old disease sub-types.
Humans can learn to recognize new categories without forgetting previously learned knowledge.
However, when state-of-the-art object detectors are fine-tuned for new tasks, they often fail on the previously trained tasks --- a problem called catastrophic forgetting \citep{goodfellow2013empirical, mccloskey1989catastrophic}.
Figure \ref{fig:catastrophic forgetting} shows an example of this problem on the PASCAL VOC dataset \citep{everingham2010pascal}.
The normal training shown in Figure \ref{fig:catastrophic forgetting} is the conventional way to make a detector work well on all tasks  --- this requires the model to be trained on labeled data from both the old and new tasks.
Unfortunately, this retraining procedure is both time-consuming and computationally expensive.
This method also requires access to all of the data for all tasks which is quite impractical in many real-life applications due to a variety of reasons.
The old training data may be inaccessible as it may have been lost or corrupted, or perhaps it is simply too large, or there may be licensing or distribution issues.

\begin{figure}[tbp]
\centering
\includegraphics[width=8.8cm, keepaspectratio]{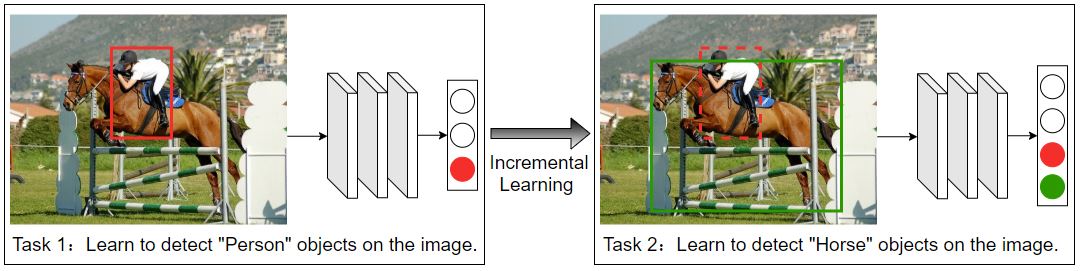}
\caption{An example of the missing annotation problem for incremental object detection \citep{shmelkov2017incremental}.
	The model is trained to incrementally learn two tasks: detect human objects and horse objects.
	First, we label all human objects in the training images for task 1 and use them to train the model.
	Then, we label all horse objects in the training images for task 2 and use them to train the model.
	If the training images for task 2 also contain human objects, the labels for them are not provided.}
\label{fig:missing annotation}
\end{figure}

In contrast to image classification where the input image only contains one class of objects, an image for object detection can contain multiple classes of objects.
Under incremental object detection scenarios, the classes of objects in one image can come from both the new task and the old tasks.
In practice, in incremental training of a previously trained detector for a specific new task, only annotations for the new object are provided and other objects are not annotated which may lead to missing annotations for the previously learned objects.
Even if all of the old data is available for normal training, these data need to be re-annotated to contain labels about all classes for all incremental learning tasks learned so far.
Figure \ref{fig:missing annotation} shows an example of the missing annotation problem.
Therefore, compared to incremental classification, incremental detection is a more challenging task, since it not only needs to solve catastrophic forgetting, but also the missing annotations for the old classes in the new data.

To bridge the performance gap between catastrophic forgetting and normal full dataset training, \citet{shmelkov2017incremental} proposed an incremental object detector using knowledge distillation \citep{hinton2015distilling}.
Their method is based on the largely superseded Fast RCNN \citep{girshick2015fast} detector which uses an external fixed proposal generator rather than a CNN, so training is not end-to-end.
To avoid missing annotation for region proposals, \citet{shmelkov2017incremental} deliberately chose the external fixed proposal generator of Fast RCNN, so that the proposals would be agnostic to the object categories.
The more recent Faster RCNN \citep{ren2015faster}  uses a trainable Region Proposal Network (RPN) to boost both accuracy and speed.
The RPN-based methods are expected to be fragile to incremental learning, because the unlabeled old class objects are treated as background during retraining of the RPN detector which may adversely affect the RPN proposals on the old classes.
In other words, if the retrained RPN is no longer able to detect the old classes by outputting proposals, there is simply no possibility of classifying them.

To address this challenge of incremental learning for RPN-based detectors, we first analyze the capability of RPN on the missing annotation problem for incremental detection.
Then, we propose an incremental framework, Faster ILOD, using multi-network adaptive distillation to improve the performance.

\textbf{The contributions} of this paper are as follows:
\begin{itemize}
\item We find that due to its unique anchor selection scheme, in the incremental detection scenario, RPN has the capability to tolerate missing annotations for old class objects to some extent.
\item Multi-network adaptive distillation is designed to further boost the accuracy of the proposed incremental object detector.
\item Using Faster RCNN \citep{ren2015faster} as the fundamental network, we demonstrate the superior performance of our model on both the PASCAL VOC \citep{everingham2010pascal} and COCO \citep{lin2014microsoft} datasets under several incremental detection settings.
\item Our framework is generic and can be applied to any object detectors using the RPN.
\end{itemize}

\section{Problem Formulation}
Incremental learning for object detection consists of $S$ incremental steps.
In each incremental step, only the batch of training data for the new classes ($C_n$) is accessible.
Given an object detection model that is previously trained using images from certain old classes ($C_o$), incremental object detection is the task of retraining the model to maintain the detection of the old classes ($C_o$) whilst detecting the new classes ($C_n$).
We refer to the original model as the old model (teacher model) and the retrained model as the new model (student model).
In multi-step incremental detection scenarios ($S>1$), for each step, all categories that were trained during any previous steps are called the old classes.
In this paper, we follow the protocol used by \citet{shmelkov2017incremental}.
More specifically, in this paper, we target the challenging real-life incremental detection scenarios as follows:
\begin{itemize}
\item In each incremental training step, only training data for the new classes is available; no representative data exemplars of the old classes from previous incremental steps are available.
\item Objects from the old classes may appear in the training images of the new detection tasks; however, the annotations for these old class objects are not provided.
\item The retrained detector should have the capability to detect objects from both the new classes and the old classes trained in all previous incremental steps.
\end{itemize}

\section{Related Work}
Our work focuses on applying Knowledge Distillation (KD) to RPN-based object detectors to improve both speed and accuracy in incremental scenarios.
In this section, we introduce the background of KD followed by a discussion about its application to incremental learning scenarios.

\subsection{Knowledge Distillation}
\label{sec: Knowledge Distillation}
KD was first introduced by \citet{hinton2015distilling} for classification model compression.
Model compression transfers the knowledge learned from a high performance cumbersome source model to a small target model.
The intuition behind KD is that the relative probabilities of incorrect answers can reveal the potential relations between different categories.
For example, in handwritten digit recognition, $7$ is more likely to be confused with $1$ than $8$.
Thus, during model compression, it is advantageous to train the target model by outputs from the source model instead of ground truth labels.
\citet{romero2014fitnets} proposed hint learning to improve the performance of model compression which distills information from feature maps of the source model.
\citet{chen2017learning} adopted both the distillation method of \cite{hinton2015distilling} and hint learning of \cite{romero2014fitnets} to detection model compression.
\citet{heo2019comprehensive} proposed a pre-ReLU feature distillation method to improve the distillation quality for model compression.

\subsection{KD based Incremental Learning Method}
As the KD method has the capability to transfer the knowledge of one model to another model, it has become one of the most commonly used tools for incremental learning.
When applying KD to incremental learning, the old model output for new data is combined with its ground truth information to train the new model.
We first discuss related methods for incremental classification followed by methods for incremental detection.

\citet{li2017learning} first applied KD to incremental learning and built an incremental classifier called LwF.
The LwF method does not require any old data to be stored and uses KD as an additional regularization term on the loss function to force the new model to follow the behavior of the old model on old tasks.
\citet{zhou2019m2kd} proposed a multi-model distillation method called M2KD which directly matches the category outputs of the current classification model with those of the corresponding old models.
Mask based pruning is used to compress the old models in M2KD.
\citet{rebuffi2017icarl} introduced a KD-based incremental classification method called iCaRL.
iCaRL stores some old data by selecting representative exemplars from each of the old classes based on herding.
The stored old exemplars and new data are combined to train the new model.
However, as only limited exemplars are stored, there is a prediction bias towards the new classes due to data size imbalance between the old and new classes.
\citet{castro2018end} kept all final classification layers during incremental learning for distillation to alleviate this data imbalance.
\citet{wu2019large} proposed using a few balanced old and new data batches to train additional two-parameter offset for the model output to remove the bias.

In contrast to classification, research on incremental object detection is quite limited in the literature.
\citet{shmelkov2017incremental} adapted the incremental classification method LwF of \citep{li2017learning}, and proposed an incremental object detector where no old data is available.
In this paper, we call their method Incremental Learning Object Detector (ILOD).
As ILOD is based on the Fast RCNN \citep{girshick2015fast} detector, it uses an external proposal generator such as EdgeBox \citep{zitnick2014edge} or MCG \citep{arbelaez2014multiscale} to generate region proposals.
\citet{shmelkov2017incremental} deliberately chose the external fixed proposal generator of Fast RCNN to ensure that proposals would be agnostic to object categories.
In our experiments, we show that our proposed method can perform well on more the efficient RPN-based detectors such as Faster RCNN \citep{ren2015faster}.

\citet{Hao2019AnEA} proposed an end-to-end incremental object detector.
In their experiments, they divided the data classes into multiple class groups and trained their model to incrementally learn the class groups.
For both training and testing of each class group, they specifically ignored all images that contain objects from multiple class groups.
This process artificially ensures that the training images for the new classes do not contain any old objects and thus avoids the missing annotation problem.
However, in real-life applications, there is a high likelihood that the input image may contain objects from both the old and new classes.

\citet{ChenLi2019ANKD} also had a similar motivation and proposed a distillation method for incremental object detection.
However, their proposed method is only evaluated on the VOC dataset over three settings without comparison to the state-of-the-art method.
Moreover, the experimental results are not clearly presented with specific accuracy in two settings.
\citet{li2019rilod} proposed a one-stage incremental object detector based on RetinaNet \citep{lin2017focal}.
In their experiments, they did not mention how they handle the annotations for old classes on new data and they only performed one-step incremental detection on the VOC dataset.

In contrast to the previous works, we target designing a high performance incremental object detector for real-life applications where new task images may also contain objects from the old classes, but the annotations for the old classes are not present.
In addition, we perform experiments for both one-step and multi-step incremental learning on two detection benchmark datasets  --- PASCAL VOC \citep{everingham2010pascal} and COCO \citep{lin2014microsoft}.

\section{Evaluation of Robustness of RPN to Incremental Object Detection}
For supervised learning, correct annotation is very important to guarantee the performance of the trained model.
Thus, some previous literature \citep{shmelkov2017incremental, Hao2019AnEA} assumed that during incremental training, the missing annotations for the old class objects would adversely affect RPN performance.
Based on this assumption, researchers have adopted several methods to avoid this problem.
\citet{shmelkov2017incremental} used the fixed external proposal generator of Fast RCNN to acquire proposals agnostic to categories.
\citet{Hao2019AnEA} carefully chose the training data to avoid old class objects appearing in new class data.

Before designing our own framework, we first evaluate how the RPN network behaves during incremental learning.
To that end, we applied the KD method in ILOD \citep{shmelkov2017incremental} to the Faster RCNN \citep{ren2015faster} detector and followed the same training strategy to train the model.
The training strategy is described in detail in Section \ref{Experiments}.
In all of our experiments, ILOD applied to Faster RCNN outperforms the original ILOD method due largely to the underlying superiority of Faster RCNN.
More specifically, according to Table \ref{tab:one_step_addition}, for one-step incremental experiments on the VOC dataset, under adding 1, 5 and 10 classes protocol, ILOD applied to Faster RCNN outperforms the original ILOD by 0.45\%, 3.63\% and 0.11\%, respectively.

According to Table \ref{tab:coco_one_step}, for one-step incremental experiments on the COCO dataset, under adding 5, 10 and 40 classes protocol, ILOD applied to Faster RCNN outperforms original ILOD by 4.1\%, 3.3\% and 3.0\%, respectively at 0.5 IoU.
According to Figure \ref{fig:fig_voc}, for multi-step incremental experiments on the VOC dataset, under each time adding 1, 2 and 5 classes protocol, ILOD applied to Faster RCNN outperforms original ILOD by an average 1.32\%, 4.11\% and 0.79\%, respectively.
According to Figure \ref{fig:fig_coco_75_5}, for multi-step incremental experiments on the COCO dataset, under each time adding 1 class protocol, ILOD applied to Faster RCNN outperforms the original ILOD by an average 3.2\% at 0.5 IoU.
In summary, unlike what was assumed by \citet{shmelkov2017incremental} and \citet{Hao2019AnEA}, we see that, the performance of Faster RCNN is not greatly affected in the case of incremental learning where the annotations of old classes are not provided in the new data.

\begin{figure}[tbp]
\centering
\includegraphics[width=8.8cm, keepaspectratio]{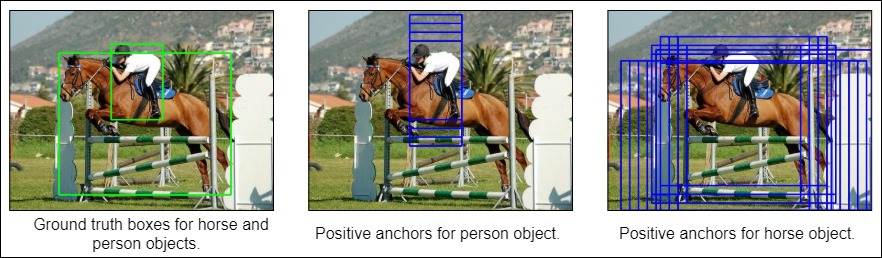}
\caption{An example of the RPN network working principle.}
\label{fig:rpn explain}
\end{figure}

Surprisingly, the RPN network is relatively robust towards missing annotations.
This behavior has also been observed by \cite{wu2018soft}.
In their experiments, after dropping 30\% of the annotations, the performance of Faster RCNN only decreases by 5\% \citep{wu2018soft}.
The task of RPN is to take the features of an image as input and output a set of category agnostic possible object proposals.
To that end, the RPN network randomly samples 256 anchors in an image to train a binary classifier, where the sampled positive and negative anchors have a ratio of up to 1:1 \citep{ren2015faster}.
An anchor is labeled as positive if it has the highest IoU overlap with a ground truth bounding box or the IoU overlap with a ground truth box is higher than a threshold (e.g. 0.7).
An anchor is labeled as negative if the IoU overlap with any ground truth box is lower than a threshold (e.g. 0.3).
Anchors that are neither positive nor negative do not contribute to RPN training.
During incremental learning, the ground truth information for the old class objects are not provided.
However, during RPN training, the method randomly samples 256 anchors from many thousands of anchors.
This means that the risk of a selected negative anchor bounding a well localized old category object is quite low.
Note further that a large proportion of the new class images may not contain any old class objects, so their annotations will be perfectly correct.
It appears that the main problem of missing annotation is that the number of positive samples used to train the RPN is reduced, since the positive anchors for the old classes are missing.

Figure \ref{fig:rpn explain} shows an example to demonstrate the working principle for the RPN network.
Three ratios (0.5, 1, 2) and five scales (2, 4, 8, 16, 32) are used for anchor generation.
The anchor stride is set to 16.
The example image has image size of 480 $\times$ 364 pixels, so 10350 anchors are generated.
After filtering repeated and excessively small anchors, 8874 anchors are available to be used.
With annotations for both the horse and person objects being provided, there are 26 positive anchors and 7307 negative anchors.
Assume the task is to incrementally learn to detect horse (new class), and the person category is regarded as the old class.
With only ground truth annotation for the horse object (new class) being provided, there are 22 positive (horse) anchors and 7476 negative anchors.
Within these 7476 negative anchors, 4 positive anchors containing the person object (old class) are wrongly regarded as negative anchors.
The false negative rate is just 0.05\%.
On the other hand, assume the task is to incrementally learn to detect a person (new class), and the horse category is regarded as the old class.
With only ground truth annotation for the person object (new class) being provided, there are 4 positive (person) anchors and 8705 negative anchors.
Within these 8705 negative anchors, 22 positive anchors containing the horse object (old class) are wrongly regarded as negative anchors.
The false negative rate is now 0.25\%.

Offsetting the adverse effects of missing annotation in the ILOD method, although distillation is only applied at the final outputs, the gradients of the distillation loss back-propagate through the entire network and will tend to encourage both the RPN and feature extractor to recognize old classes.
This helps explain why RPN training is not catastrophically affected by the missing annotations --- at least not over the range of our experiments, such as one and several-step incremental settings.

\section{Faster ILOD for Robust Incremental Object Detection}
\begin{figure*}[htbp]
\centering
\subfigure[Under the add one new class at a time protocol.]{\begin{minipage}[t]{0.34\linewidth}
		\centering
		\includegraphics[width=\textwidth,keepaspectratio]{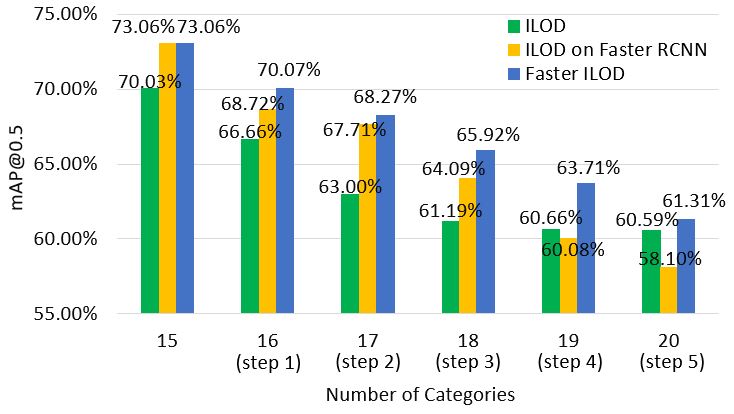}
		\label{fig:fig_voc_15_5}		
\end{minipage}}
\quad
\subfigure[Under the add two new classes at a time protocol.]{\begin{minipage}[t]{0.33\linewidth}
		\centering
		\includegraphics[width=\textwidth,keepaspectratio]{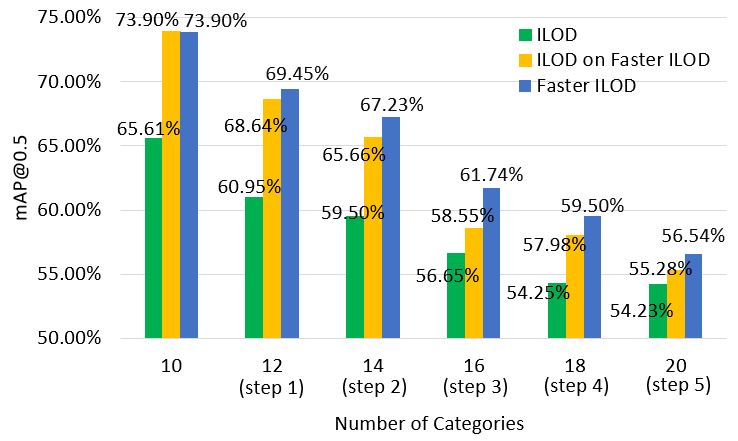}	
		\label{fig:fig_voc_10_10}		
\end{minipage}}
\quad
\subfigure[Under the add five new classes at a time protocol.]{\begin{minipage}[t]{0.275\linewidth}
		\centering
		\includegraphics[width=\textwidth,keepaspectratio]{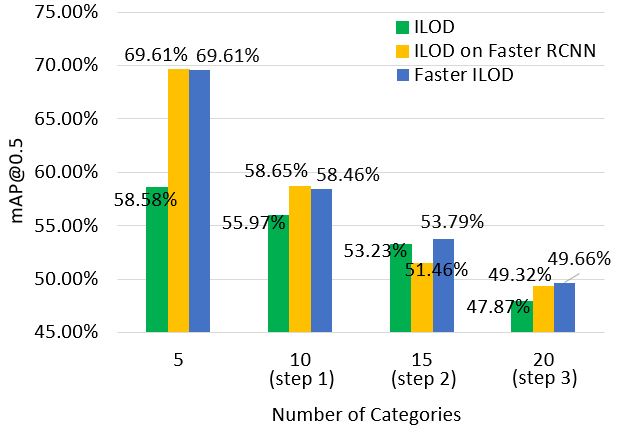}	
		\label{fig:fig_voc_5_15}			
\end{minipage}}%
\caption{Overall mAP accuracy performance for ILOD, ILOD applied to Faster RCNN and Faster ILOD on the VOC dataset under different multi-step incremental scenarios.}
\label{fig:fig_voc}
\end{figure*}

\begin{figure*}[tbp]
\centering	
\subfigure[mAP at 0.5 IoU.]{\begin{minipage}[t]{0.345\linewidth}
		\centering
		\includegraphics[width=\textwidth,keepaspectratio]{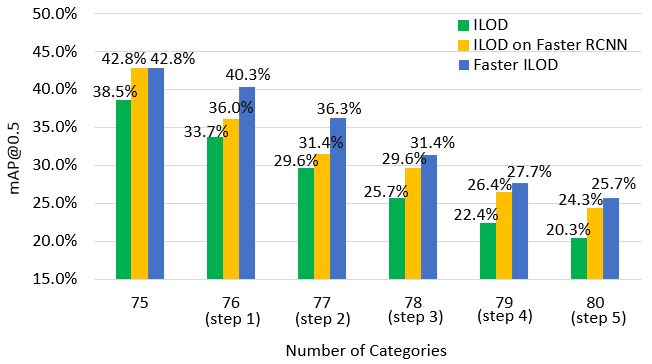}
		\label{fig:fig_coco_75_5_0.5}		
\end{minipage}}
\quad
\subfigure[mAP weighted across different IoU from 0.5 to 0.95.]{\begin{minipage}[t]{0.345\linewidth}
		\centering
		\includegraphics[width=\textwidth,keepaspectratio]{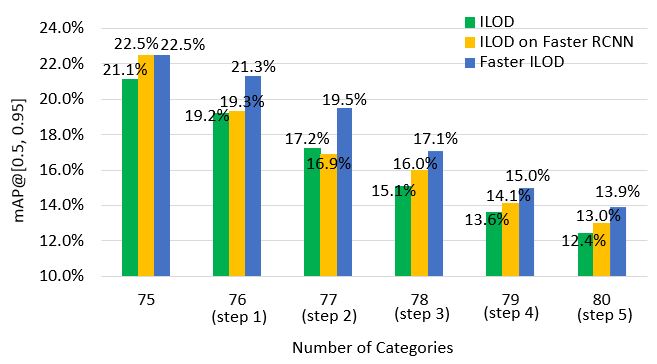}	
		\label{fig:fig_coco_75_5_avg}		
\end{minipage}}%
\caption{Overall mAP accuracy performance for ILOD, ILOD applied to Faster RCNN and Faster ILOD on the COCO dataset under the add one new class at a time protocol.}
\label{fig:fig_coco_75_5}
\end{figure*}

There remains an accuracy gap between the ILOD method applied to RPN-based detectors and full data training.
For example, according to Figure \ref{fig:fig_voc}, for the VOC dataset, the full data training result is 69.50\% but under multi-step incremental learning, all the results for ILOD applied to Faster RCNN are less than 60\% accuracy.
According to Figure \ref{fig:fig_coco_75_5}, for the COCO dataset, the full data training result is 42.71\% at 0.5 IoU but under multi-step incremental setting, the result for ILOD applied to Faster RCNN is only 24\% at 0.5 IoU.
In this section, we propose a novel multi-network adaptive distillation method to further narrow the gap.
We first discuss the backbone network used for our model and then discuss each component of our proposed method.

\begin{figure}[tbp]
\centering
\includegraphics[width=7cm, keepaspectratio]{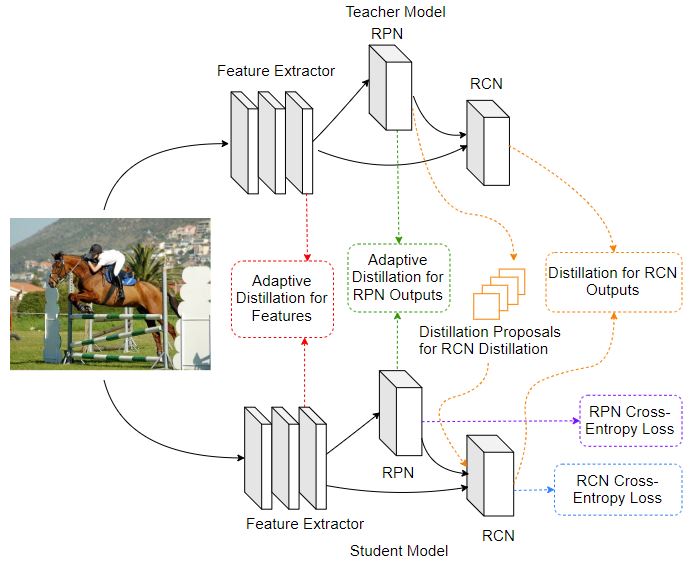}
\caption{Framework of our proposed method to perform incremental object detection on Faster RCNN \citep{ren2015faster}:
	Adaptive distillation on the feature maps and RPN outputs works together with a conventional distillation component on the final outputs to preserve previous knowledge and alleviate missing annotation problem for the old classes on new data.}
\label{fig:framework}
\end{figure}

\subsection{Object Detection Network}
Our proposed method for incremental object detection is illustrated by Figure \ref{fig:framework}.
It comprises two models: a teacher model ($N_{te}$) and a student model ($N_{st}$).
The teacher model is a frozen copy of the original detector which detects objects from the old categories ($C_{te}$ = $C_o$).
The student model is the adapted model that needs to be trained to detect objects from both the old and new categories ($C_{st}$ = $C_o$ $\cup$ $C_n$).
It is also initially a copy of the original detector but the number of outputs in the last layer is increased to predict for the additional new classes.
We use Faster RCNN \citep{ren2015faster} as our backbone network.
Faster RCNN is a two-stage end-to-end object detector which consists of three parts: (1) A Convolutional Neural Network (CNN) based feature extractor to provide features; (2) a Region Proposal Network (RPN) to produce regions of interest (RoIs); (3) a class-level classification and bounding box regression network (RCN) to generate the final prediction for each of the proposals from RPN \citep{chen2017learning}.
In order to create a high performance incremental object detector, it is important to properly account for all three components.

\subsection{Multi-Network Adaptive Distillation}
\label{sec: adaptive distillation}
To make a model remember what it learned before, similar to ILOD \citep{shmelkov2017incremental}, we adapt knowledge distillation.
But unlike ILOD which only performs one-step distillation at the final outputs, we perform multi-network distillation on the feature maps, RPN and RCN outputs.
In addition, knowledge distillation is originally developed for model compression which only requires the preservation of learned knowledge.
Incremental learning requires not only maintaining learned knowledge, but also learning new knowledge from the new classes.
Thus, directly applying distillation loss to force the student model to follow the behavior of the teacher model will simply prevent new data learning.
To solve this problem, we propose adaptive distillation which uses the teacher model outputs as a lower bound to adaptively distill old knowledge.

\textbf{Feature Distillation}: The desired feature extractor needs to provide features that are effective for both old and new categories.
To build the desired feature extractor, we utilize normalized adaptive distillation with a $\mathcal L_1$ loss.
Specifically, all feature maps are normalized to obtain the corresponding zero-mean feature maps, $\tilde{F}_{te}$ and $\tilde{F}_{st}$, from the teacher model and student model, respectively.
For each activation in the feature map, we then check its value from the student model ($\tilde{f}_{st}\in\tilde{F}_{st}$) with the corresponding value from the teacher model ($\tilde{f}_{te}\in\tilde{F}_{te}$).
If the teacher's activation, $\tilde{f}_{te}$, has a higher value, a loss is generated to force the student model to increase its value for this input, since this activation is important for the old classes.
On the other hand, if the student's activation, $\tilde{f}_{st}$, has a higher value, the loss is zero since this activation is likely important for the new classes.
The feature distillation loss is:

\begin{small}
\begin{equation}
\mathcal L_{F\_Dist}=\frac{1}{\mathcal M}\sum
\begin{cases}
\begin{Vmatrix}\tilde{f}_{te} - \tilde{f}_{st}\end{Vmatrix}_{1}\textrm{ }, & \mbox{if }\tilde{f}_{te} > \tilde{f}_{st}\\
0\textrm{ }, & \mbox{otherwise}
\end{cases}
\label{eq: feature dist}
\end{equation}
\end{small}

\noindent where $te$ and $st$ subscripts refer to teacher and student networks respectively and $\mathcal{M}$ is the total number of activation values in the feature map.

\textbf{RPN Distillation}: The desired RPN needs to provide proposals for objects from both new and old classes.
Similar to feature distillation loss, we use the teacher model RPN output as a lower bound to force the student model to choose anchors according to both the training data from the new classes and the teacher model RPN output.
In addition, the bounding box regression can provide incorrect values since the real valued regression output is unbounded and partially trained only at positive anchors.
Inspired by the distillation method used for detection model compression by \cite{chen2017learning}, we use a threshold, $\mathcal{T}$, to control regression.
In our experiments, empirically we set $\mathcal{T} = 0.1$.
For RPN distillation, $\mathcal L_{2}$ loss is used.
Suppose $\mathcal N$ is the total number of anchors, $q$ is the RPN classification output, and $r$ is the RPN bounding box regression output.
The RPN distillation loss is:

\begin{small}
\begin{equation}
\mathcal L_{RPN\_Dist}=\frac{1}{\mathcal N}\sum
\begin{cases}
\begin{Vmatrix} q_{te} - q_{st}\end{Vmatrix}_{2}^{2} +\beta \begin{Vmatrix}r_{te} - r_{st}\end{Vmatrix}_{2}^{2}\textrm{, }&\mbox{if }q_{te} > q_{st}\\
0\textrm{ },&\mbox{otherwise}
\end{cases}
\label{eq: RPN dist}
\end{equation}
\noindent where
\begin{equation}
\nonumber
\beta=\begin{cases} 1\textrm{ }, &\mbox{if }q_{te} > (q_{st} + \mathcal{T})\\
0\textrm{ }, &\mbox{otherwise.}
\end{cases}
\end{equation}
\end{small}

\textbf{RCN Distillation}: The desired RCN needs to predict each RoI for both old and new classes in an unbiased manner.
We follow the method in ILOD \citep{shmelkov2017incremental} to perform RCN distillation.
More specifically, for each image, we randomly choose 64 out of 128 RoIs with the smallest background score according to the RPN output from the teacher model.
Then these proposals are fed into the RCN of the student model and the final outputs of the teacher model are used as targets for the old classes.
The outputs of student model on the new classes are not included in the RCN distillation.
For each RoI classification output, $p$, we subtract the mean over the class dimension to get the zero-mean classification result, $\tilde{p}$.
We use $\mathcal L_2$ loss for the distillation.
Let $\mathcal K$ be the total number of sampled RoIs, $C_o$ be the number of old classes including background, and $t$ be the bounding box regression result.
The RCN distillation loss is then written as:

\begin{small}
\begin{equation}
\mathcal L_{RCN\_Dist}=\frac{1}{\mathcal K \times C_o} \sum \left[\begin{Vmatrix}\tilde{p}_{te} - \tilde{p}_{st}\end{Vmatrix}_{2}^{2}+\begin{Vmatrix}t_{te} - t_{st}\end{Vmatrix}_{2}^{2}\right]
\label{eq: RCN dist}
\end{equation}
\end{small}

\textbf{Total Loss Function}: The overall loss ($\mathcal L_{total}$) will be the weighted summation of the standard Faster R-CNN loss \citep{ren2015faster}, feature distillation loss \eqref{eq: feature dist}, RPN distillation loss \eqref{eq: RPN dist}, and RCN distillation loss \eqref{eq: RCN dist}.
Hyper-parameters $\lambda_{1}$, $\lambda_{2}$ and $\lambda_{3}$ help to balance each loss term, and are empirically set to 1.

\begin{small}
\begin{equation}
\mathcal{L}_{total}=\mathcal{L}_{RCNN} +\lambda_{1} \mathcal{L}_{F\_Dist} +\lambda_{2} \mathcal{L}_{RPN\_Dist} + \lambda_{3} \mathcal{L}_{RCN\_Dist}
\label{eq: total dist}
\end{equation}
\end{small}

\section{Experiments}
\label{Experiments}

\begin{table*}[tpb]
\centering
\scriptsize{
	\caption{Average mAP for one-step incremental experiments on the VOC dataset.
		$N_{te}$(a-b) is the first-step normal training for categories a to b.
		+$N_{st}$(c) is the incremental learning result after adding c new categories.
	}
	\begin{tabular}{|l|c|c|c|c|c|c|c|}
		\hline
		& $N_{te}$(1-19) & $+N_{st}$(1) & $N_{te}$(1-15) & $+N_{st}$(5) & $N_{te}$(1-10) & $+N_{st}$(10) & $N_{te}$(1-20) \\
		\hline
		ILOD~\citep{shmelkov2017incremental} & 68.60\% & 67.27\% & 70.03\% & 62.72\% & 65.61\% & 61.03\% & 69.50\% \\
		\hline
		ILOD applied to Faster RCNN & \multirow{3}*{70.10\%} & 67.72\% & \multirow{3}*{73.06\%} & 66.35\% & \multirow{3}*{73.90\%} & 61.14\% & \multirow{3}*{70.52\%} \\
		\cline{1-1} \cline{3-3} \cline{5-5} \cline{7-7}
		Faster ILOD w/o feature distillation &  & 68.32\% &  & 67.73\% &  & 61.78\% &  \\
		\cline{1-1} \cline{3-3} \cline{5-5} \cline{7-7}
		Faster ILOD &  & \bf{68.56\%} &  & \bf{67.94\%} &  & \bf{62.16\%} &  \\
		\hline
	\end{tabular}	
	\label{tab:one_step_addition}}
~\\	
\centering
\scriptsize{
	\caption{
		Average mAP for one-step incremental experiments on the COCO dataset.
		The detection accuracy is evaluated by mAP at 0.5 IoU (mAP@.5) and mAP weighted across different IoU from 0.5 to 0.95 (mAP@[0.5, 0.95]).
	}
	\begin{tabular}{|l|c|c|c|c|c|c|c|}
		\hline
		& $N_{te}$(1-75) & $+N_{st}$(5) & $N_{te}$(1-70) & $+N_{st}$(10) & $N_{te}$(1-40) & $+N_{st}$(40) & $N_{te}$(1-80) \\
		\hline
		ILOD~\citep{shmelkov2017incremental} (mAP@.5) & 38.52\% & 33.80\% & 38.39\% & 34.26\% & 40.01\% & 36.11\% & 38.22\% \\
		\hline
		ILOD applied to Faster RCNN (mAP@.5) & \multirow{2}*{42.78\%} & 37.87\% & \multirow{2}*{42.95\%} & 37.59\% & \multirow{2}*{46.98\%} & 39.10\% & \multirow{2}*{42.71\%} \\
		\cline{1-1} \cline{3-3} \cline{5-5} \cline{7-7}
		Faster ILOD (mAP@.5) &  & \bf{39.60\%} &  & \bf{39.87\%} &  & \bf{40.10\%} &  \\
		\hline
		\hline
		ILOD~\citep{shmelkov2017incremental} (mAP@[.5, .95]) & 21.10\% & 19.19\% & 21.73\% & 19.52\% & 22.69\% & 19.83\% & 21.21\% \\
		\hline
		ILOD applied to Faster RCNN (mAP@[.5, .95]) & \multirow{2}*{22.45\%} & 20.02\% & \multirow{2}*{22.90\%} & 19.90\% & \multirow{2}*{24.43\%} & 20.21\% & \multirow{2}*{22.52\%} \\
		\cline{1-1} \cline{3-3} \cline{5-5} \cline{7-7}
		Faster ILOD (mAP@[.5, .95]) &  & \bf{21.03\%} &  & \bf{21.28\%} &  & \bf{20.64\%} &  \\
		\hline
	\end{tabular}
	\label{tab:coco_one_step}}
~\\	
\scriptsize{
	\caption{Average mAP for one-step incremental experiments on the VOC dataset for adding different categories.
		$N_{te}$(1-19) is the first-step normal training for 19 categories. +$N_{st}$(c) is the incremental learning result for category c.
	}
	\begin{tabular}{|l|c|c|c|c|c|c|c|}
		\hline
		& $N_{te}$(1-19) & $+N_{st}$(TV) & $N_{te}$(1-19) & $+N_{st}$(aeroplane) & $N_{te}$(1-19) & $+N_{st}$(person) & $N_{te}$(1-20) \\
		\hline
		Amount of training data & 4931 & 256 & 4806 & 238 & 4485 & 2008 & 5011 \\
		\hline
		\tabincell{l}{Number of old categories \\ occurring in the training data} & - & 9 & - & 6 & - & 19 & - \\
		\hline
		Percentage of missing annotation & - & 60.40\% & - & 22.22\% & - & 40.87\% & - \\
		\hline
		ILOD applied to Faster RCNN & \multirow{2}*{70.10\%} & 67.72\% & \multirow{2}*{70.32\%} & 67.62\% & \multirow{2}*{70.08\%} & 65.79\% & \multirow{2}*{70.52\%} \\
		\cline{1-1} \cline{3-3} \cline{5-5} \cline{7-7}
		Faster ILOD &  & \bf{68.56\%} &  & \bf{68.92\%} &  & \bf{69.57\%} &  \\
		\hline
	\end{tabular}	
	\label{tab:one_step_diff_category}
}
\end{table*}

We call our proposed method Faster ILOD as it is designed to work with Faster RCNN.
In this section, we compared our Faster ILOD method with the original ILOD method as well as ILOD applied to Faster RCNN.

\subsection{Dataset and Evaluation Metric}
In our experiments, two detection benchmark datasets are used, PASCAL VOC 2007 \citep{everingham2010pascal} and COCO 2014 \citep{lin2014microsoft}.
VOC 2007 comprises 10k images of 20 object categories --- 5k for training and 5k for testing.
COCO 2014 comprises 164k images of 80 object categories --- 83k for training, 40k for validation and 41k for testing.
For the evaluation metric, we use mean average precision (mAP) at 0.5 Intersection over Union (IoU) for both datasets and also use mAP weighted across different IoU from 0.5 to 0.95 for COCO.
To validate our method, we have investigated several incremental settings for these two datasets, such as one-step and multi-step addition.
The sequence of categories is arranged according to the category names in alphabetical order.

\subsection{Implementation Details}
The results for ILOD \citep{shmelkov2017incremental} are generated using their public implementation.
Edge-Boxes \citep{zitnick2014edge} is used to generate the external proposals.
Faster ILOD and ILOD applied to Faster RCNN are implemented using PyTorch. 
For a fair comparison of our approach with ILOD \citep{shmelkov2017incremental}, we use the same backbone network (ResNet-50 \citep{he2016deep}) and similar training strategy mentioned in their paper.
In the first step of training, we set the learning rate to 0.001, decaying to 0.0001 after 30k iterations, and momentum is set to 0.9.
The network is trained using 40k iterations for VOC and 400k iterations for COCO.
In the following incremental steps, learning rate is set to 0.0001.
The network is trained using 5k-10k iterations when only one class is added and the same number of iterations as the first step if multiple classes are added at once.

\subsection{Experiments on VOC Dataset}
Table \ref{tab:one_step_addition} shows the results for one-step incremental settings when the number of new classes equals 1, 5 and 10, respectively.
In all three settings, Faster ILOD is more accurate than both ILOD and ILOD applied to Faster RCNN.
Under these settings, we also performed experiments on Faster ILOD without adaptive feature distillation to prove the efficiency of each part of our method.
Compared to the experimental results for multi-step increments, we see the improvement is not significant in the one-step settings.
This is likely because one-step increments require a small amount of fine-tuning on the old model and the catastrophic forgetting and missing annotation problems might not be significant, which provides little room for improvement.
When we retrain the old model in the multiple incremental steps, the build-up of detection errors due to catastrophic forgetting and missing annotation are approximately exponential on the old classes. So this is a more difficult scenario.

We have also investigated the results under multi-step incremental scenarios.
Figure \ref{fig:fig_voc_15_5} shows the performance of Faster ILOD, ILOD and ILOD applied to Faster RCNN, when first training with 15 classes followed by the addition of 1 class for 5 steps.
Observing from Figure \ref{fig:fig_voc_15_5}, under the add one new class at a time protocol, Faster ILOD outperforms ILOD and ILOD applied to Faster RCNN in each incremental step and the average performance gain is 3.44\% and 2.12\% respectively.
Figure \ref{fig:fig_voc_10_10} shows the performance of three models under the condition of first training with 10 classes followed by addition of 2 classes for 5 times.
Under this incremental setting, Faster ILOD also performs best for all five incremental steps and the average accuracy improvement is 5.78\% towards ILOD and 1.67\% towards ILOD applied to Faster RCNN.
Figure \ref{fig:fig_voc_5_15} shows the performance of three models under the condition of first training with 5 classes followed by addition of 5 classes for 3 times.
Under this incremental scenario, Faster ILOD outperforms ILOD applied to Faster RCNN except the first step and always has better accuracy than ILOD.
The average accuracy increase is 1.61\% towards ILOD and 0.83\% towards ILOD applied to Faster RCNN.

\subsection{Experiments on COCO Dataset}
For our experiments on the COCO dataset, we use train set and valminusminival set as our training data and minival set as our testing data.
Table \ref{tab:coco_one_step} shows the results under one-step incremental settings, where the number of new classes is 5, 10 and 40, respectively.
Figure \ref{fig:fig_coco_75_5} shows the results for multi-step incremental detection under the add one new class at a time protocol.
In both scenarios, Faster ILOD provides the best detection accuracy.
In particular, under multi-step incremental detection shown in Figure \ref{fig:fig_coco_75_5}, Faster ILOD outperforms ILOD and ILOD applied to Faster RCNN in all steps and has an average gain of 5.92\% and 2.74\% (1.88\% and 1.53\%) respectively at 0.5 IoU (weighted across different IoU from 0.5 to 0.95).

\subsection{Discussions}
To explore how learning different categories affects the model performance on the same incremental setting, under the one-step incremental learning scenario, we performed the experiments on adding different categories on the VOC dataset.
These experiments indicate that our method can always achieve improvements in different learning sequence.
Table \ref{tab:one_step_diff_category} shows the results for incrementally learning the new class (TV and aeroplane).
It also shows the results for incrementally learning the more challenging person category.
Comparing with other categories, incrementally learning person category is more challenging since ‘person’ objects often appear with many other category objects in one image.
In VOC training data, there are 2008 out of 5011 images containing person objects.
Within the 2008 training images, there are 3641 other objects (40.87\% missing annotation) which belong to 19 old categories.
As for person category, the amount of training data is large, the model is trained for 40,000 instead of 10,000 iterations for convergence.
For all three different categories, our proposed Faster ILOD outperforms ILOD applied to Faster RCNN.
According to Table \ref{tab:one_step_diff_category}, the improvement of our adaptive distillation method for person category is higher than aeroplane and TV categories which have fewer missing annotations (55 and 408, respectively) and fewer training data (238 and 256, respectively).
Although the missing annotation percentage for TV category is high (60.40\%), it only comes from 9 old categories.

In summary, for different categories, they will face different difficulty levels for incremental learning.
We conjecture that the difficulty levels are related to three items:
(1) the amount of training data;
(2) the number of old categories co-occurring in the new training data;
(3) the percentage of missing annotations in the training data.
Therefore, the challenge of incremental learning is different under different incremental learning scenarios which leads to different accuracy improvements of our method.

\subsection{Evaluations on Detection Speed}
As the original ILOD code is built in Tensorflow, to fairly compare the detection speeds for ILOD and Faster ILOD, we rebuilt ILOD on Pytorch.
All experiments were performed on an NVIDIA Tesla V100 GPU.
Average detection time per image of ILOD and Faster ILOD with ResNet-50 \citep{he2016deep} on the VOC dataset is 1396.66 ms and 109.52 ms respectively.
As ILOD relies on an external proposal generator to acquire proposals, the inference speed of ILOD is about 13 times slower than Faster ILOD due to the fixed proposal network.

\section{Conclusion}

In this paper, we find that the RPN network is relatively robust towards missing annotations for old classes on incremental object detection and then propose a novel end-to-end framework, Faster ILOD.
By adaptively distilling the old information in multi-networks, the proposed method aims to preserve the capabilities of the detector on old classes with limited affect towards the learning on new classes.
Our method shows superior results on the PASCAL VOC and COCO datasets and outperforms the state-of-the-art incremental detector \citep{shmelkov2017incremental} by a large margin in most cases.

\section*{Acknowledgments}
This research was funded by the Australian Government through the Australian Research Council and Sullivan Nicolaides Pathology under Linkage Project LP160101797.

\bibliographystyle{model2-names}
\bibliography{refs}

\end{document}